\documentclass[lettersize,journal]{IEEEtran}
\usepackage{amsmath,amsfonts}
\usepackage{algorithmic}
\usepackage{array}
\usepackage[caption=false,font=normalsize,labelfont=sf,textfont=sf]{subfig}
\usepackage{textcomp}
\usepackage{stfloats}
\usepackage{url}
\usepackage{verbatim}
\usepackage{graphicx}
\usepackage{booktabs}
\def\BibTeX{{\rm B\kern-.05em{\sc i\kern-.025em b}\kern-.08em
    T\kern-.1667em\lower.7ex\hbox{E}\kern-.125emX}}
\usepackage{balance}
\usepackage{makecell}
\usepackage{multirow}
\usepackage{graphicx}
\usepackage{CJKutf8} \usepackage{orcidlink}

\begin{document}
\begin{CJK*}{UTF8}{gbsn}

\title{Dual-Stream Simultaneous Translation via 2D Grid Attention}
\author{Yu~Pu\orcidlink{0009-0008-9414-2013} and Wei-Qiang~Zhang\orcidlink{0000-0003-3841-1959},~\IEEEmembership{Senior Member,~IEEE}\thanks{This work has been submitted to the IEEE for possible publication. Copyright may be transferred without notice, after which this version may no longer be accessible.}
\thanks{This work was supported by the National Natural Science Foundation of China under Grant No. 62276153. \emph{(Corresponding author: Wei-Qiang Zhang.)}}
\thanks{Y.~Pu and W.-Q.~Zhang are with the Department of Electronic Engineering, Tsinghua University, Beijing 100084, China. W.-Q. Zhang is also with the Institute for Embodied Intelligence and Robotics, Tsinghua University, Beijing 100084, China. (e-mail: puy23@mails.tsinghua.edu.cn; wqzhang@tsinghua.edu.cn).}}

\markboth{IEEE/ACM Transactions on Audio, Speech, and Language Processing}{Pu and Zhang: Dual-Stream Simultaneous Translation via 2D Grid Attention}

\maketitle

\begin{abstract}
Simultaneous machine translation must generate target tokens before the source input is complete. Existing approaches address this through post-hoc read-write policies, leaving the attention mechanism unaware of bidirectional stream dependencies. We propose a dual-stream attention framework that represents source and target streams as a two-dimensional grid of hidden states and models their interaction through four structurally distinct attention types merged via joint QK Softmax normalization. Two approximations---broadcast and Hadamard---reduce the per-layer complexity from $O(X^2Y{+}XY^2)$ to $O(X^2{+}Y^2{+}XY)$ with provably decaying error. Training uses a self-guided loop: a per-cell loss heatmap drives dynamic-programming path recovery, which generates read/write decision supervision labels without external alignment. An incremental KV cache with anchored rotary position embeddings enables efficient streaming inference. On Chinese-to-English simultaneous translation, the proposed model outperforms the Wait-$k$ baseline by $+5.66$ BLEURT and $+10.36$ COMET at comparable latency, and surpasses the non-streaming reference on COMET at a fraction of the response delay.
\end{abstract}


\section{Introduction}
\label{sec:intro}

\IEEEPARstart{S}{imultaneous} machine translation (SimulMT) targets a qualitatively different operating regime from offline translation: a target-language token must be emitted \emph{before} the full source sentence is \cite{wang2024simultaneous}. This constraint is not merely a latency budget; it demands that the model continuously reason about an incomplete, still-arriving source stream while producing a coherent, causally valid output stream \cite{fu2025llms}. The resulting \textbf{latency-quality trade-off}---emit too early and translation degrades; wait too long and real-time utility is lost---is the central challenge of the field \cite{wang2025conversational}.

The dominant paradigm addresses this trade-off through \textbf{post-hoc waiting policies} layered on top of standard offline architectures \cite{papi2022does}. Fixed policies such as Wait-$k$~\cite{elbayad2020efficient} prescribe a deterministic schedule: read $k$ source tokens, then alternate one read with one write. Adaptive policies~\cite{arivazhagan2019massively,dalvi2018incremental} learn content-dependent schedules via reinforcement learning or multi-path training. These approaches are practical and effective, yet they share a structural limitation: the underlying attention mechanism remains \emph{unaware} of the bidirectional dependency between streams \cite{zhao2023adaptive}. The decoder attends to a frozen source prefix and generates autoregressively, with no mechanism for the source representations to respond to what has already been generated, or for the output representations to dynamically integrate newly arriving source tokens. The read-write schedule is thus imposed on top of the computation, rather than being woven into it.

We address this gap by modeling stream dependencies directly at the attention level, treating simultaneous translation as a \textbf{bidirectional interaction} between two co-evolving streams. Instead of a one-dimensional input sequence and a separately unrolled decoder, we represent the joint state at any point $(x,y)$---having consumed $x$ source tokens and emitted $y$ target tokens---as a node in a \textbf{2D grid} of hidden states, indexed by both source position and output time-step. Within this grid, we identify four structurally distinct attention patterns and instantiate each with a dedicated computation:
\begin{itemize}
  \item \textbf{I$\to$I} (source self-attention, causal along $x$): captures left-to-right dependencies within the source stream.
  \item \textbf{O$\to$O} (target self-attention, causal along $y$): maintains autoregressive coherence in the output stream.
  \item \textbf{I$\leftarrow$O} (output-to-input cross-attention, prefix along $y$): allows source representations to be updated by the partial output generated so far.
  \item \textbf{O$\leftarrow$I} (input-to-output cross-attention, prefix along $x$): allows output representations to attend to the source prefix available at each state.
\end{itemize}
Because each output node $(x,y)$ receives contributions from both its own-stream and cross-stream attention, we derive a \textbf{joint QK Softmax} formulation that merges the unnormalized statistics from O$\to$O and O$\leftarrow$I in a single numerically stable pass.

Full grid computation is quadratic in both $X$ and $Y$, which is prohibitive at scale. We therefore develop two \textbf{low-complexity approximations}: (1) a \emph{broadcast approximation} that computes I$\leftarrow$O attention only at $y=0$ and replicates the result along the $y$ axis, reducing its cost from $O(X^2 Y)$ to $O(X^2)$; and (2) a \emph{Hadamard approximation} that replaces the cross-position dot product in I$\leftarrow$O with an element-wise product at the same position, cutting the remaining cross-stream cost to $O(XY)$. We derive a formal error bound showing that the resulting approximation error decays as $O(1/\sqrt{y})$ as the output prefix grows, providing theoretical justification for the approximation quality.

Training exploits the 2D grid structure to derive \textbf{self-guided supervision} without external alignment labels. We compute a \emph{loss heatmap} $\mathcal{L}(x,y)$ from grid-level language-model log-likelihoods, then recover the \emph{optimal monotone path} $\pi^*$ through this heatmap via dynamic programming. The path determines an emission mask that labels each grid state as ``should emit'' or ``should wait'', providing direct binary supervision for a lightweight decision head. Crucially, as translation quality improves during training, the heatmap reshapes, yielding better paths and therefore better decision labels---a self-reinforcing loop that tightens both objectives simultaneously.

At inference time, incremental decoding requires maintaining eight categories of KV-cache tensors per layer to avoid redundant recomputation. We design an \textbf{incremental KV cache} that exploits the causal and prefix masking structure of each attention type, and pair it with an \textbf{anchored RoPE} strategy that assigns output-stream positions starting from $X_{\text{total}}$, eliminating position-embedding conflicts between the two streams.

Experiments on Chinese-to-English simultaneous translation demonstrate that at a comparable output speed (FRL $\approx 3.3$ vs.\ $3.0$), our model achieves $+5.66$ BLEURT and $+10.36$ COMET over the Wait-$k$ ($k{=}3$) baseline, while also outperforming the non-streaming reference on COMET at the most permissive latency setting ($\theta{=}0.9$, COMET $64.73$ vs.\ $63.85$).

The contributions of this paper are as follows:
\begin{enumerate}
  \item A \textbf{2D grid attention framework} with four structurally distinct stream interaction types, together with a joint QK Softmax for numerically stable merging of cross-stream attention statistics.
  \item Two \textbf{efficient approximations}---broadcast and Hadamard---with a formal error analysis showing $O(1/\sqrt{y})$ decay.
  \item A \textbf{self-guided co-training} procedure based on loss heatmap, DP path recovery, and an emission mask, requiring no external alignment supervision.
  \item An \textbf{incremental inference} design with a structured KV cache and anchored RoPE that supports low-latency streaming decoding.
\end{enumerate}

The remainder of the paper is organized as follows. Section~\ref{sec:related} reviews related work. Section~\ref{sec:bistream-attn} presents the duplex attention model and joint QK Softmax. Section~\ref{sec:approx} develops the low-complexity approximations and error analysis. Section~\ref{sec:training} describes the training and inference algorithms. Section~\ref{sec:experiments} reports experimental results, and Section~\ref{sec:conclusion} concludes.

\section{Related Work}
\label{sec:related}

\subsection{Simultaneous Translation Strategies and Policies}

Simultaneous translation (SimulMT) requires generating target-language output before the source input is complete, departing fundamentally from the offline ``read-then-translate'' paradigm~\cite{xia2023speculative}. The central challenge is the \textbf{latency-quality trade-off}: early emission risks quality degradation from insufficient context, while excessive waiting violates real-time constraints \cite{zhang2021universal}. Task variants span text-to-text (SimulMT), speech-to-text (SimulST), and speech-to-speech (SimulS2S) translation, with SimulMT serving as the methodological foundation for the others \cite{ren2020simulspeech}.

Prior work has pursued this challenge through two broad families of read-write policies. Fixed policies prescribe a deterministic schedule regardless of content. The Wait-$k$ strategy~\cite{elbayad2020efficient,arivazhagan2020re,han2021monotonic,chen2021improving} is the most widely adopted: the decoder first reads $k$ source tokens and then alternates between consuming one source token and emitting one target token. The prefix-to-prefix framework~\cite{xia2023speculative,lin2023leapt,kano2022simultaneous} takes a complementary approach, training models to translate from source prefixes alone; within this framework, Ma et al.~\cite{ma2019stacl} proposed STACL, which incorporates an anticipation mechanism to improve prefix-conditioned generation under limited context.

Adaptive policies, by contrast, learn content-dependent schedules. Arivazhagan et al.~\cite{arivazhagan2019massively} employ deep reinforcement learning to directly optimize a read-write scheduling policy, rewarding favorable latency-quality outcomes. Xia et al.~\cite{xia2023speculative} introduce speculative decoding, where a lightweight predictor proposes candidate tokens verified by the main model, reducing effective latency without quality degradation. Dalvi et al.~\cite{dalvi2018incremental} propose incremental training with multi-path optimization, exposing the model to diverse read-write trajectories to improve robustness across latency budgets. Curriculum-based approaches~\cite{zhang2019curriculum} further ease training by gradually decreasing the latency tolerance across epochs.

\subsection{Simultaneous Translation Model Training and Evaluation}

Architecturally, most simultaneous systems extend the standard autoregressive Transformer \cite{guo2024decoder,zhang2021future,ma2023non}. Policy-integrated variants introduce dedicated read/write decision heads alongside token logits, decoupling translation quality from scheduling \cite{zheng2020simultaneous,zheng2019simpler,zhang2020learning}. Monotonic attention mechanisms constrain the source-target alignment to be strictly left-to-right, enabling online decoding without look-ahead \cite{liu2021cross,lee2021simultaneous}. Dual-stream designs separate input reading from output generation into distinct submodules, making bidirectional dependencies explicit within the computation graph \cite{xia2024dual}. Compared to offline NMT, simultaneous systems commonly adopt multi-path training~\cite{dalvi2018incremental, bao2023multi}, randomly sampling read-write trajectories at each training step so that the model learns to handle a range of prefix lengths simultaneously. Knowledge distillation from an offline teacher further stabilizes this process: the offline model provides soft targets under full source context, guiding the simultaneous model toward more robust prefix-conditioned predictions \cite{wang2023better,yang2024translate,wan2024dual}. Prefix-alignment augmentation and curriculum scheduling~\cite{zhang2019curriculum} are complementary strategies that similarly mitigate the information scarcity inherent to simultaneous decoding.

For evaluation, the field employs latency-specific metrics alongside standard quality measures such as BLEU~\cite{papineni2002bleu}. Average Lagging (AL)~\cite{elbayad2020efficient} quantifies the mean temporal offset between source and target token generation; Differentiable AL~\cite{dalvi2018incremental} makes this quantity amenable to end-to-end gradient-based optimization. First Token Latency is particularly salient for user perception in real-time settings. Most evaluations are conducted on WMT \cite{specia2021findings,zerva2022findings,blain2023findings} and IWSLT \cite{ansari2020findings,agarwal2023findings,agostinelli2025findings} benchmarks, reporting results along the quality-latency Pareto frontier under varying latency budgets.

\section{Duplex Attention Modeling}
\label{sec:bistream-attn}

\subsection{Core Challenge of Simultaneous Translation}

Traditional interaction systems follow a ``non-streaming'' paradigm: speakers take turns occupying the channel, and the system begins output only after receiving the complete input \cite{wang2022progress}. This paradigm is fundamentally inadequate for simultaneous interpretation, where the interpreter must continuously produce the target-language output while simultaneously listening to the source \cite{guo2025streamuni}. Under non-streaming modeling, the system either introduces unacceptable translation latency by waiting for the input to end, or sacrifices translation quality by outputting prematurely.

The core challenge of simultaneous translation modeling is: how to explicitly represent the \textit{bidirectional dependency between the input stream and the output stream} within the same neural network \cite{dong2022learning}. Let the input sequence be $\mathbf{i} = (i_1, i_2, \ldots, i_X)$ and the output sequence be $\mathbf{o} = (o_1, o_2, \ldots, o_Y)$. At any state $(x, y)$ (having seen $x$ inputs and generated $y$ outputs), the model must simultaneously satisfy:
\begin{enumerate}
  \item \textbf{Output causality}: $o_{y+1}$ can only depend on $i_1,\ldots,i_x$ and $o_1,\ldots,o_y$, and cannot ``see'' future inputs;
  \item \textbf{Input awareness}: output generation should dynamically adapt as new inputs arrive, rather than being deferred until the input ends.
\end{enumerate}

Existing work typically inserts waiting strategies (e.g., CIF \cite{dong2020cif}, Wait-$k$) as post-processing, without directly modeling these dependencies at the attention level of the language model. Our dual-stream model explicitly introduces cross-stream dependencies inside the attention mechanism of a large language model, simultaneously satisfying both constraints within a unified framework.

\subsection{Inference State Machine}

We formalize simultaneous inference as a deterministic state machine. Define state $s = (x, y)$ as ``$x$ input tokens have been exposed to the model and $y$ output tokens have been emitted externally.'' The state machine supports two actions:
\begin{itemize}
  \item \textbf{EMIT}: generate the $(y+1)$-th output token; state transitions $(x,y) \to (x, y+1)$;
  \item \textbf{WAIT}: expose the $(x+1)$-th input token; state transitions $(x,y) \to (x+1, y)$.
\end{itemize}
At state $(x,y)$, the model must output: (1) the probability distribution $p(o_{y+1} \mid i_{1:x}, o_{1:y})$ over the next output token; (2) the EMIT/WAIT decision probability $p(\mathrm{EMIT} \mid i_{1:x}, o_{1:y})$.

Table~\ref{tab:notation} lists the principal symbols used throughout this and the following sections.

\begin{table*}[t]
  \caption{Principal Notation}
  \label{tab:notation}
  \centering
  \begin{tabular}{cll}
    \toprule
    \textbf{Symbol} & \textbf{Meaning} & \textbf{Typical Value} \\
    \midrule
    $B$ & Batch size & --- \\
    $X$ & Input sequence length & --- \\
    $Y$ & Output sequence length & --- \\
    $D$ & Model hidden dimension & 896 \\
    $H$ & Number of attention heads & 14 \\
    $H_k$ & Number of Key/Value heads (GQA) & 2 \\
    $H_d$ & Per-head dimension; $H_d = D / H$ & 64 \\
    $L$ & Number of Transformer layers & 24 \\
    $s$ & Attention scaling factor; $s = H_d^{-1/2}$ & --- \\
    $\odot$ & Element-wise (Hadamard) product & --- \\
    $\mathbf{I}^{(l)}$ & Input-stream hidden states at layer $l$; $\mathbf{I}^{(l)} \in \mathbb{R}^{B \times X \times Y \times D}$ & --- \\
    $\mathbf{O}^{(l)}$ & Output-stream hidden states at layer $l$; $\mathbf{O}^{(l)} \in \mathbb{R}^{B \times X \times Y \times D}$ & --- \\
    $\tilde{\mathbf{I}}^{(l)}, \tilde{\mathbf{O}}^{(l)}$ & LayerNorm-normalized inputs to attention at layer $l$ & --- \\
    $W_Q, W_K, W_V$ & Shared QKV projection matrices; $W_Q \in \mathbb{R}^{D \times HH_d}$, $W_K, W_V \in \mathbb{R}^{D \times H_k H_d}$ & --- \\
    $W_O$ & Output projection matrix; $W_O \in \mathbb{R}^{H H_d \times D}$ & --- \\
    $\mathcal{L}(x,y)$ & Grid loss at state $(x,y)$ & --- \\
    $\pi^*$ & Optimal monotone path from $(0,0)$ to $(X,Y)$ & --- \\
    $\lambda$ & Latency tolerance hyperparameter in path score & 0.1 \\
    $\theta$ & EMIT decision threshold at inference time & --- \\
    \bottomrule
  \end{tabular}
\end{table*}

\subsection{2D Grid Hidden State Representation}

Standard autoregressive language models represent each token in a sequence as a one-dimensional vector. We extend dual-stream modeling to a two-dimensional grid. For layer $l$, define:
\begin{equation}
  \mathbf{I}^{(l)} \in \mathbb{R}^{B \times X \times Y \times D}, \quad
  \mathbf{O}^{(l)} \in \mathbb{R}^{B \times X \times Y \times D},
  \label{eq:grid-def}
\end{equation}
where $\mathbf{I}^{(l)}_{x,y}$ denotes the hidden state of the $x$-th input token at layer $l$ under the context of having seen $x$ inputs and $y$ outputs, and $\mathbf{O}^{(l)}_{x,y}$ denotes the corresponding hidden state for the $y$-th output token. The same token thus has different hidden state representations at different states $(x,y)$, reflecting the variation of its semantics under different contexts.

At layer 0 (the embedding layer), the grid is initialized by broadcasting token embeddings:
\begin{equation}
  \mathbf{I}^{(0)}_{x,y} = \mathrm{Embed}(i_x), \quad
  \mathbf{O}^{(0)}_{x,y} = \mathrm{Embed}(o_y).
  \label{eq:grid-init}
\end{equation}
Note that $\mathbf{I}^{(0)}_{x,y}$ is independent of $y$, and $\mathbf{O}^{(0)}_{x,y}$ is independent of $x$. In subsequent Transformer layers, cross-stream attention introduces mutual dependencies along both dimensions.

\begin{figure}[htbp]
  \centering
  \includegraphics[width=1.0\linewidth]{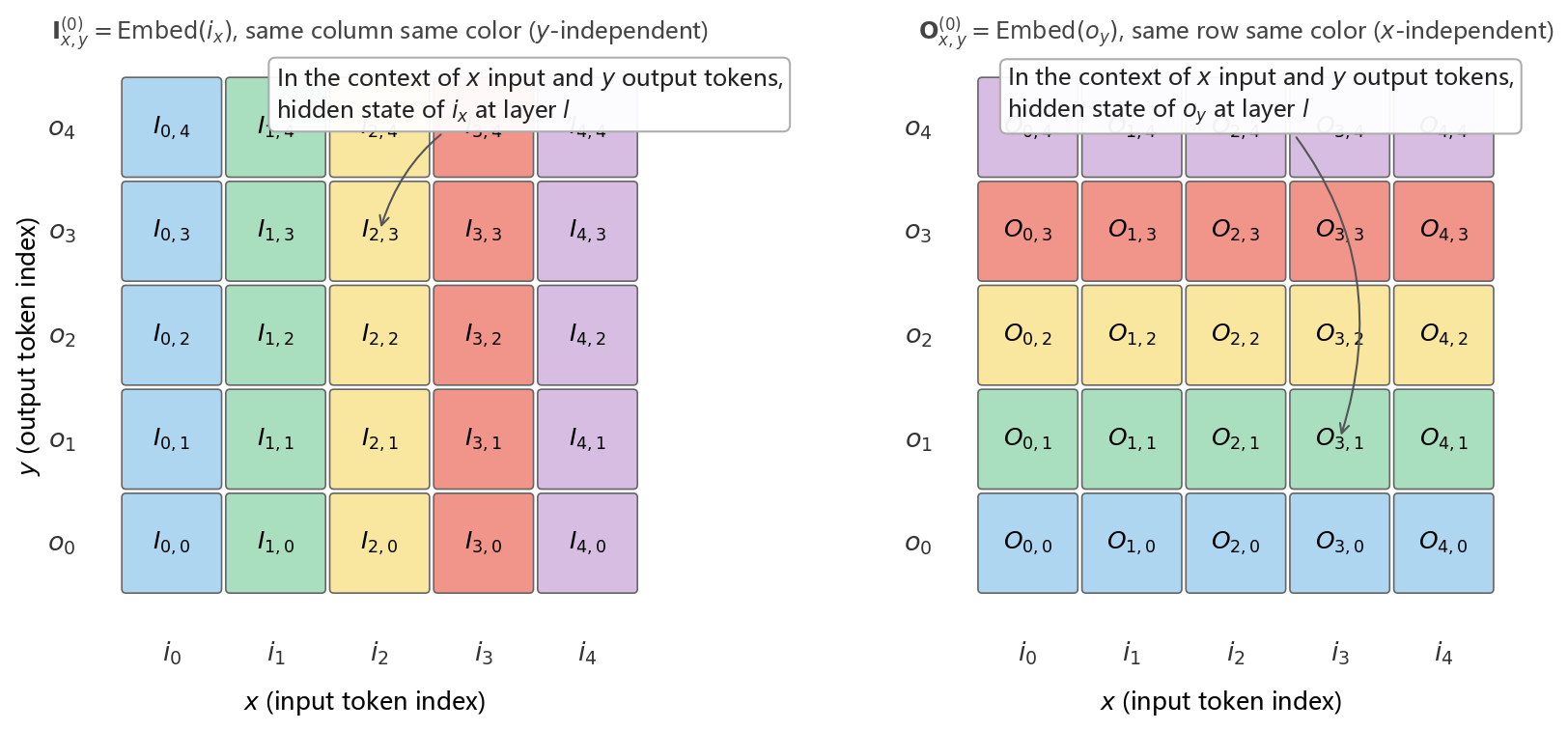}
  \caption{Illustration of the 2D grid hidden state representation. Each grid cell $(x,y)$ maintains hidden states for both the input stream $\mathbf{I}^{(l)}_{x,y}$ and the output stream $\mathbf{O}^{(l)}_{x,y}$. The horizontal axis corresponds to input positions and the vertical axis to output positions.}
  \label{fig:grid-repr}
\end{figure}

For positional encoding, we assign input tokens positions $0, 1, \ldots, X-1$ and output tokens positions $X, X+1, \ldots, X+Y-1$ within the same Rotary Position Embedding (RoPE) sequence:
\begin{equation}
  \mathbf{Q}_I[x,y,h], \mathbf{K}_I[x,y,h] \xleftarrow{\mathrm{RoPE}(x)} \text{position } x,
\end{equation}
\begin{equation}
  \mathbf{Q}_O[x,y,h], \mathbf{K}_O[x,y,h] \xleftarrow{\mathrm{RoPE}(X+y)} \text{position } X+y.
\end{equation}
This design ensures that intra-stream relative positions are consistent with standard autoregressive models, enabling full reuse of pre-trained positional priors.

\subsection{Four Types of Attention}

In each Transformer layer \cite{vaswani2017attention}, attention contexts are computed separately for the input and output streams, followed by residual connections and MLP to obtain the next-layer hidden states. The dual-stream attention comprises four types, as summarized in Table~\ref{tab:attn-types}.

\begin{table}[htbp]
  \caption{Four Types of Duplex Attention and Their Properties}
  \label{tab:attn-types}
  \centering
  \begin{tabular}{lllll}
    \toprule
    Type & Query & Key/Value & Causal Constraint \\
    \midrule
    I$\to$I & $\mathbf{Q}_I$ & $\mathbf{K}_I, \mathbf{V}_I$ & Causal in $x$ ($x' \le x$) \\
    O$\to$O & $\mathbf{Q}_O$ & $\mathbf{K}_O, \mathbf{V}_O$ & Causal in $y$ ($y' \le y$) \\
    I$\leftarrow$O & $\mathbf{Q}_I$ & $\mathbf{K}_O, \mathbf{V}_O$ & Prefix in $y$ ($y' \le y$) \\
    O$\leftarrow$I & $\mathbf{Q}_O$ & $\mathbf{K}_I, \mathbf{V}_I$ & Prefix in $x$ ($x' \le x$) \\
    \bottomrule
  \end{tabular}
\end{table}

\begin{figure}[htbp]
  \centering
  \includegraphics[width=1.0\linewidth]{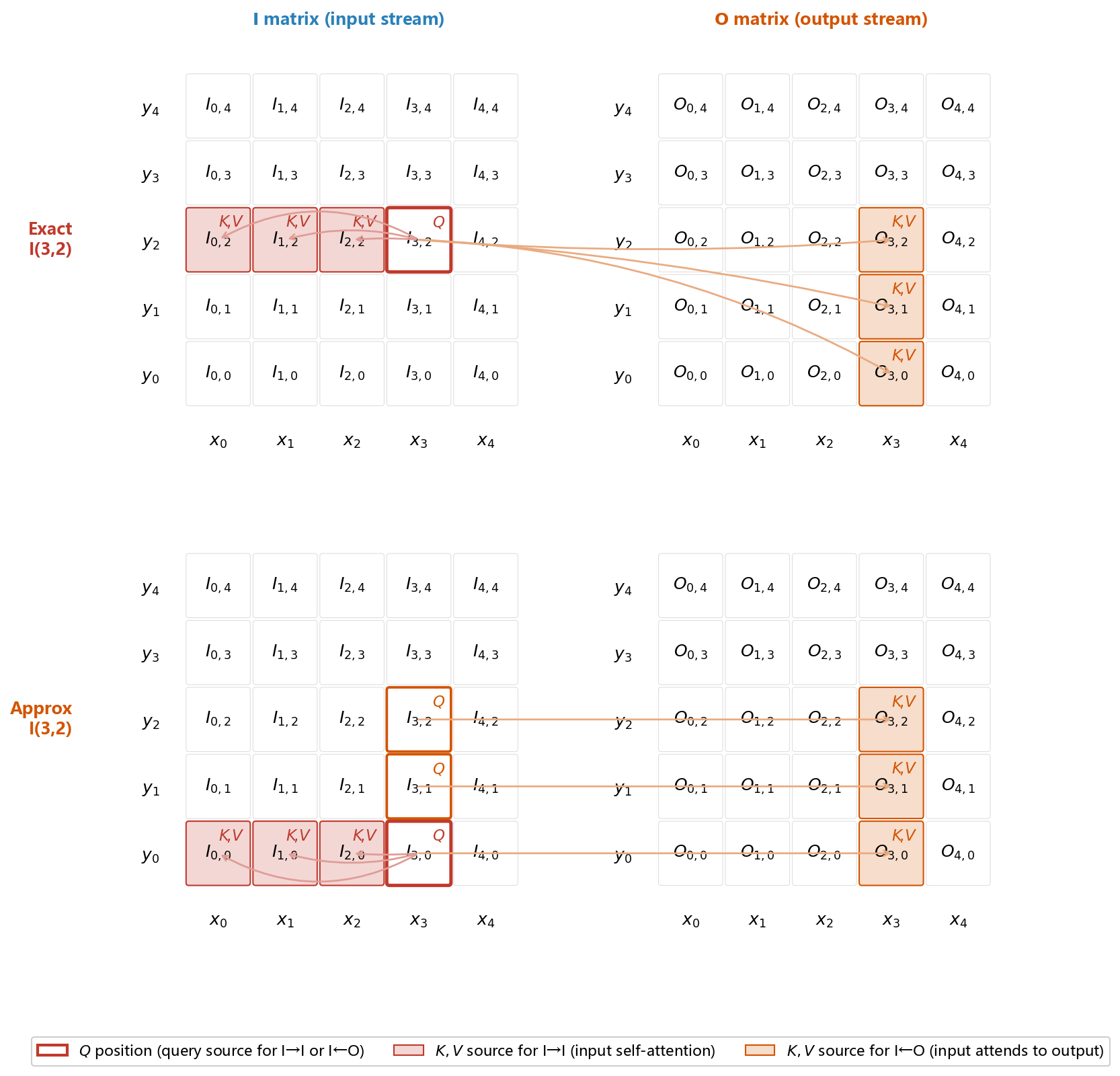}
  \caption{Grid illustration of input stream attention. Left: I$\to$I self-attention, where each cell attends causally along $x$ and results are broadcast along $y$. Right: I$\leftarrow$O cross-attention, where each input cell aggregates from the output prefix $y' \le y$.}
  \label{fig:attn-types-i}
\end{figure}

\begin{figure}[htbp]
  \centering
  \includegraphics[width=1.0\linewidth]{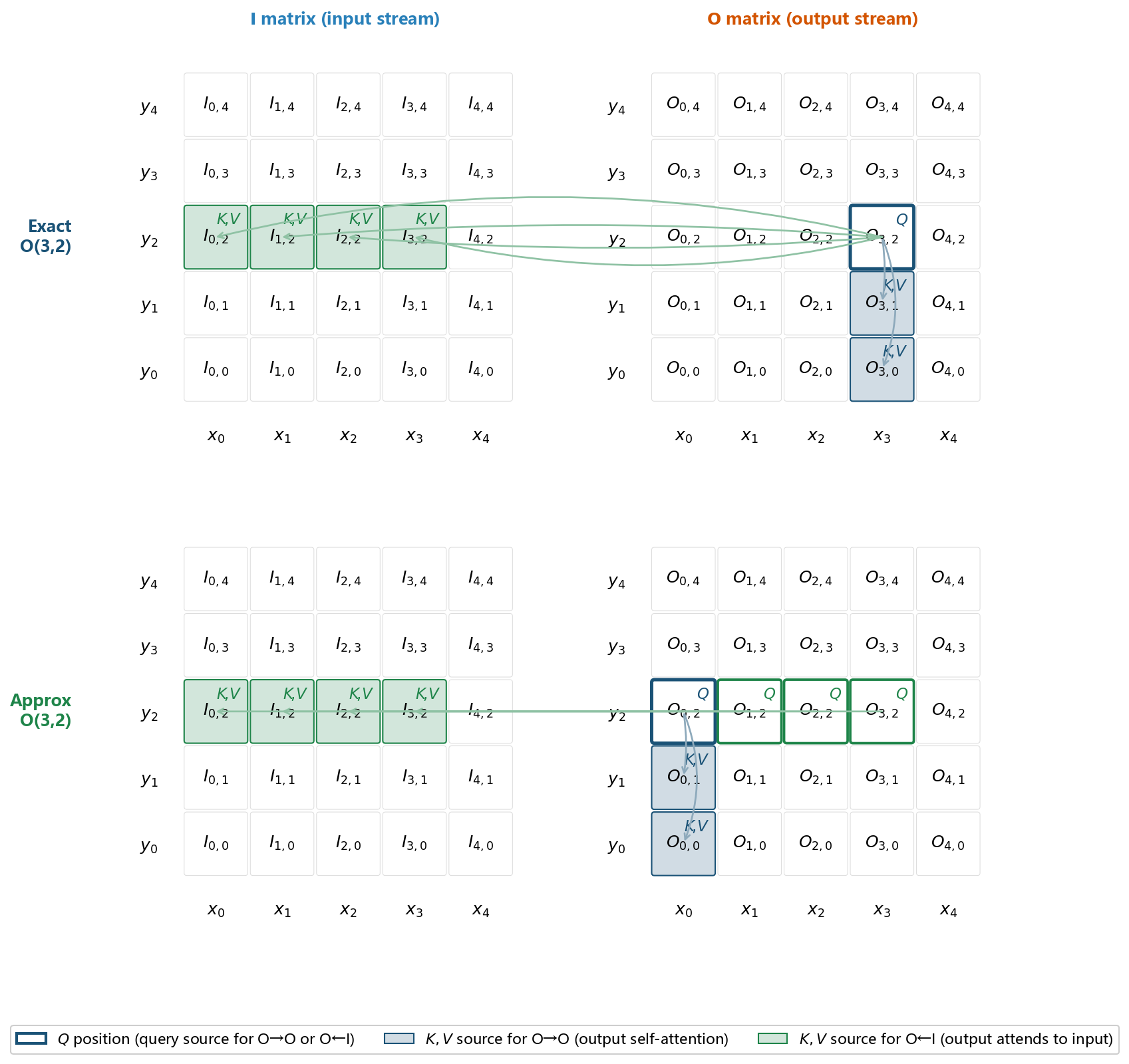}
  \caption{Grid illustration of output stream attention. Left: O$\to$O self-attention, broadcast along $x$. Right: O$\leftarrow$I cross-attention, where each output cell aggregates from the input prefix $x' \le x$.}
  \label{fig:attn-types-o}
\end{figure}

For layer $l$, both streams first apply LayerNorm, and their query, key, and value tensors are projected via shared weight matrices $W_Q \in \mathbb{R}^{D \times HH_d}$ and $W_K, W_V \in \mathbb{R}^{D \times H_k H_d}$ (with GQA heads expanded to $H$):
\begin{align}
  \mathbf{Q}_I &= \tilde{\mathbf{I}}^{(l)} W_Q, \quad \mathbf{K}_I = \tilde{\mathbf{I}}^{(l)} W_K, \quad \mathbf{V}_I = \tilde{\mathbf{I}}^{(l)} W_V, \\
  \mathbf{Q}_O &= \tilde{\mathbf{O}}^{(l)} W_Q, \quad \mathbf{K}_O = \tilde{\mathbf{O}}^{(l)} W_K, \quad \mathbf{V}_O = \tilde{\mathbf{O}}^{(l)} W_V,
\end{align}
where $\tilde{\mathbf{I}}^{(l)}, \tilde{\mathbf{O}}^{(l)}$ are the LayerNorm-normalized hidden states and $s = H_d^{-1/2}$ is the attention scaling factor. Sharing $W_Q, W_K, W_V$ across both streams fully reuses pre-trained weights.

\subsubsection{I$\to$I Self-Attention (Causal along $x$)}
The input stream attends causally to its own tokens:
\begin{equation}
  a^{\mathrm{ii}}_{x,x'} = s \cdot \mathbf{Q}_I[x,h] \cdot \mathbf{K}_I[x',h]^\top, \quad x' \le x,
\end{equation}
\begin{equation}
  \mathrm{ctx}^{\mathrm{ii}}_x = \sum_{x'=0}^{x} \frac{e^{a^{\mathrm{ii}}_{x,x'}}}{\sum_{k=0}^{x} e^{a^{\mathrm{ii}}_{x,k}}} \mathbf{V}_I[x',h].
\end{equation}
Since this result depends only on $x$, it is broadcast across all output positions: $\mathrm{ctx}^{\mathrm{ii}}[x,y] = \mathrm{ctx}^{\mathrm{ii}}_x$, $\forall\,y$.

\subsubsection{O$\to$O Self-Attention (Causal along $y$)}
Symmetric to I$\to$I, the output stream attends causally to its own tokens:
\begin{equation}
  a^{\mathrm{oo}}_{y,y'} = s \cdot \mathbf{Q}_O[y,h] \cdot \mathbf{K}_O[y',h]^\top, \quad y' \le y,
\end{equation}
\begin{equation}
  \mathrm{ctx}^{\mathrm{oo}}_y = \sum_{y'=0}^{y} \frac{e^{a^{\mathrm{oo}}_{y,y'}}}{\sum_{k=0}^{y} e^{a^{\mathrm{oo}}_{y,k}}} \mathbf{V}_O[y',h],
\end{equation}
broadcast as $\mathrm{ctx}^{\mathrm{oo}}[x,y] = \mathrm{ctx}^{\mathrm{oo}}_y$, $\forall\,x$.

\subsubsection{I$\leftarrow$O Cross-Attention (Prefix along $y$)}
Each input token aggregates information from the output prefix generated so far. Using the Hadamard approximation (Section~\ref{sec:approx}), the attention score at grid cell $(x,y)$ is:
\begin{equation}
  s^{\mathrm{io}}_{x,y,h} = s \cdot \bigl(\mathbf{Q}_I[x,y,h] \odot \mathbf{K}_O[x,y,h]\bigr) \cdot \mathbf{1},
\end{equation}
and the context is a prefix Softmax weighted sum over $y' \le y$:
\begin{equation}
  \mathrm{ctx}^{\mathrm{io}}[x,y,h] = \frac{\displaystyle\sum_{y'=0}^{y} e^{s^{\mathrm{io}}_{x,y',h} - m^{\mathrm{io}}_{x,y,h}} \mathbf{V}_O[x,y',h]}{Z^{\mathrm{io}}_{x,y,h}},
\end{equation}
where $m^{\mathrm{io}}_{x,y,h} = \max_{y'\le y} s^{\mathrm{io}}_{x,y',h}$ is the running maximum and $Z^{\mathrm{io}}_{x,y,h}$ is the normalization constant.

\subsubsection{O$\leftarrow$I Cross-Attention (Prefix along $x$)}
Symmetric to I$\leftarrow$O, each output token aggregates the input prefix:
\begin{equation}
  s^{\mathrm{oi}}_{x,y,h} = s \cdot \bigl(\mathbf{Q}_O[x,y,h] \odot \mathbf{K}_I[x,y,h]\bigr) \cdot \mathbf{1},
\end{equation}
\begin{equation}
  \mathrm{ctx}^{\mathrm{oi}}[x,y,h] = \frac{\displaystyle\sum_{x'=0}^{x} e^{s^{\mathrm{oi}}_{x',y,h} - m^{\mathrm{oi}}_{x,y,h}} \mathbf{V}_I[x',y,h]}{Z^{\mathrm{oi}}_{x,y,h}}.
\end{equation}

A naive approach would combine the two attention outputs by simple addition after independent normalization: $\mathrm{ctx}_I = W_O(\mathrm{ctx}^{\mathrm{ii}} + \mathrm{ctx}^{\mathrm{io}})$. However, because each of $\mathrm{ctx}^{\mathrm{ii}}$ and $\mathrm{ctx}^{\mathrm{io}}$ is already individually Softmax-normalized, their sum always weights them equally regardless of the actual attention score magnitudes, preventing the model from learning which type of attention is more informative for a given state.

We instead merge via a \textbf{joint QK Softmax normalization}, combining the unnormalized running statistics $(m, Z, S)$ of both attention types under a shared normalization constant before applying $W_O$. For the I-stream:
\begin{align}
  m_j &= \max(m^{\mathrm{ii}}_x,\; m^{\mathrm{io}}_{x,y}), \\
  Z_j &= Z^{\mathrm{ii}}_x \cdot e^{m^{\mathrm{ii}}_x - m_j} + Z^{\mathrm{io}}_{x,y} \cdot e^{m^{\mathrm{io}}_{x,y} - m_j}, \\
  S_j &= S^{\mathrm{ii}}_x \cdot e^{m^{\mathrm{ii}}_x - m_j} + S^{\mathrm{io}}_{x,y} \cdot e^{m^{\mathrm{io}}_{x,y} - m_j}, \\
  \mathrm{ctx}_I[x,y] &= W_O\!\left(S_j / Z_j\right).
\end{align}
This is equivalent to placing all I$\to$I and I$\leftarrow$O key-value pairs into the same Softmax denominator, allowing the model to allocate attention weight between self-attention and cross-attention in a data-driven manner. The O-stream is merged symmetrically using $m^{\mathrm{oo}}_y$ and $m^{\mathrm{oi}}_{x,y}$:
\begin{align}
  m_j^O &= \max(m^{\mathrm{oo}}_y,\; m^{\mathrm{oi}}_{x,y}), \\
  Z_j^O &= Z^{\mathrm{oo}}_y \cdot e^{m^{\mathrm{oo}}_y - m_j^O} + Z^{\mathrm{oi}}_{x,y} \cdot e^{m^{\mathrm{oi}}_{x,y} - m_j^O}, \\
  \mathrm{ctx}_O[x,y] &= W_O\!\left(\bigl(S^{\mathrm{oo}}_y \cdot e^{m^{\mathrm{oo}}_y - m_j^O} + S^{\mathrm{oi}}_{x,y} \cdot e^{m^{\mathrm{oi}}_{x,y} - m_j^O}\bigr) / Z_j^O\right).
\end{align}
The attention outputs are then applied via residual connections to yield the next-layer hidden states:
\begin{align}
  \mathbf{I}^{(l)}_{x,y} &= \mathrm{MLP}\!\left(\mathbf{I}^{(l-1)}_{x,y} + \mathrm{ctx}_I[x,y]\right), \\
  \mathbf{O}^{(l)}_{x,y} &= \mathrm{MLP}\!\left(\mathbf{O}^{(l-1)}_{x,y} + \mathrm{ctx}_O[x,y]\right),
\end{align}
where MLP denotes the standard two-layer feed-forward sub-block with LayerNorm (identical in structure to a vanilla Transformer layer). Both streams share the same MLP weights, again fully reusing pre-trained parameters.

\subsection{Online Prefix Softmax Updates}
The core computation of cross-attention is a prefix Softmax weighted sum, and maintaining this sum incrementally is essential for efficient inference. For the I$\leftarrow$O cross-attention, we maintain three prefix statistics for each $(x,y,h)$:
\begin{align}
  m^{\mathrm{io}}_{x,y} &\in \mathbb{R}^{B \times X \times H}, \\
  Z^{\mathrm{io}}_{x,y} &\in \mathbb{R}^{B \times X \times H}, \\
  S^{\mathrm{io}}_{x,y} &\in \mathbb{R}^{B \times X \times H \times H_d},
\end{align}
initialized as $m^{(0)} = -\infty$, $Z^{(0)} = 0$, $S^{(0)} = \mathbf{0}$. When a new output position $y$ arrives, the update rules are:
\begin{align}
  m^{\mathrm{new}} &= \max\!\left(m^{\mathrm{old}},\; s^{\mathrm{io}}_{x,y}\right), \\
  Z^{\mathrm{new}} &= Z^{\mathrm{old}} \cdot e^{m^{\mathrm{old}} - m^{\mathrm{new}}} + e^{s^{\mathrm{io}}_{x,y} - m^{\mathrm{new}}}, \\
  S^{\mathrm{new}} &= S^{\mathrm{old}} \cdot e^{m^{\mathrm{old}} - m^{\mathrm{new}}} + e^{s^{\mathrm{io}}_{x,y} - m^{\mathrm{new}}} \cdot \mathbf{V}_O[x,y], \\
  \mathrm{ctx}^{\mathrm{io}}[x,y] &= S^{\mathrm{new}} \, / \, Z^{\mathrm{new}}.
\end{align}
This online log-sum-exp algorithm is numerically stable and analogous to the FlashAttention prefix Softmax routine~\cite{dao2022flashattention}. The O$\leftarrow$I prefix statistics are maintained symmetrically along the $x$ dimension.

\section{Low-Complexity Approximations}
\label{sec:approx}

Without approximation, the exact computation of duplex attention has total complexity $O(X^2Y + XY^2)$ per layer, which is prohibitive for long sequences. We propose two approximations that reduce the complexity to $O(X^2 + Y^2 + XY)$.

\subsection{Broadcast Approximation}

We compute I$\to$I self-attention only at column $y=0$ and broadcast the result across the entire grid:
\begin{equation}
  \mathrm{ctx}^{\mathrm{ii}}[x, y] \approx \mathrm{ctx}^{\mathrm{ii}}[x, 0], \quad \forall\, y.
  \label{eq:broadcast-approx}
\end{equation}
The symmetric approximation applies to O$\to$O. This reduces the self-attention complexity from $O(X^2Y + XY^2)$ to $O(X^2 + Y^2)$.

The approximation error stems from the $y$-dependence of the I-stream hidden state $\mathbf{I}^{(l)}[x,y]$. At initialization (layer 0), $\mathbf{I}^{(0)}[x,y] = \mathrm{Embed}(i_x)$ is exactly $y$-independent, so the error is exactly zero at the start of training and grows gradually as cross-stream attention builds up. Semantically, the intra-stream attention pattern (which input attends to which) should primarily depend on input content and relative positions, not on how many output tokens have been generated---making this approximation linguistically well-motivated.

\subsection{Hadamard Approximation}

For I$\leftarrow$O cross-attention, the exact score at grid position $(x,y)$ for attending to key at position $y'$ requires a dot product $\mathbf{Q}_I[x,y,h] \cdot \mathbf{K}_O[x,y',h]^\top$. We approximate this by replacing the query at $y$ with the query at $y'$, turning the cross-position dot product into an element-wise product at the same position:
\begin{equation}
  a^{\mathrm{io}}_{x,y,y'} \approx s \cdot \bigl(\mathbf{Q}_I[x,y',h] \odot \mathbf{K}_O[x,y',h]\bigr) \cdot \mathbf{1}.
  \label{eq:io-hadamard}
\end{equation}
This allows each grid cell $(x,y')$ to pre-compute a scalar score, enabling prefix Softmax accumulation with $O(H_d)$ cost per step, reducing cross-attention complexity to $O(XY)$.

\subsubsection{Error Analysis}
The Hadamard approximation assumes the query $\mathbf{Q}_I[x,y,h]$ is approximately constant in $y$. The per-pair score error is:
\begin{equation}
  \Delta^{\mathrm{io}}(x,y,y') = s \cdot \bigl(\mathbf{Q}_I[x,y,h] - \mathbf{Q}_I[x,y',h]\bigr) \cdot \mathbf{K}_O[x,y',h]^\top.
  \label{eq:hadamard-error}
\end{equation}
Since $\mathbf{Q}_I[x,y,h] = W_Q \tilde{\mathbf{I}}^{(l)}[x,y]$, the query variation is linearly controlled by the I-stream hidden-state perturbation $\delta\mathbf{I}^{(l)}[x,y] = \mathbf{I}^{(l)}[x,y] - \mathbf{I}^{(l)}[x,0]$, giving the error bound:
\begin{equation}
  \bigl|\Delta^{\mathrm{io}}(x,y,y')\bigr| \le s \cdot \|W_Q\|_2 \cdot \|\delta\mathbf{I}^{(l)}[x,y{-}y']\|_2 \cdot \|\mathbf{K}_O[x,y',h]\|_2.
  \label{eq:hadamard-error-bound}
\end{equation}
Both the broadcast and Hadamard errors are therefore governed by the same root cause: the amplitude of I$\leftarrow$O cross-attention. At layer 0, $\delta\mathbf{I}^{(0)} = \mathbf{0}$, so both approximations are exact at initialization and diverge only gradually as cross-stream attention builds up during training.

Furthermore, the Hadamard approximation introduces \emph{per-pair score} errors rather than direct context-vector errors. Within the prefix Softmax, individual score deviations are smoothed by the shared normalization: if the errors $\Delta^{\mathrm{io}}$ have mixed signs across positions $y'$, their net effect on the normalized context vector decays as $O(1/\sqrt{y})$ with prefix length. In practice, attention weights tend to be concentrated on a small number of positions, so errors at low-weight positions contribute negligibly to the final context.

Table~\ref{tab:complexity} summarizes the complexity before and after approximation.

\begin{table}[htbp]
  \caption{Computational Complexity Per Layer Before and After Approximation}
  \label{tab:complexity}
  \centering
  \begin{tabular}{llll}
    \toprule
    Attention Type & Exact & Approximated \\
    \midrule
    I$\to$I self-attention & $O(X^2 Y)$ & $O(X^2)$ \\
    O$\to$O self-attention & $O(X Y^2)$ & $O(Y^2)$ \\
    I$\leftarrow$O cross-attention & $O(X Y^2)$ & $O(XY)$ \\
    O$\leftarrow$I cross-attention & $O(X^2 Y)$ & $O(XY)$ \\
    \midrule
    Total & $O(X^2Y + XY^2)$ & $O(X^2 + Y^2 + XY)$ \\
    \bottomrule
  \end{tabular}
\end{table}

For typical sentence lengths $X = Y = 30$, this reduces the per-layer operation count by approximately $20\times$, bringing duplex attention to the same order as standard single-stream autoregressive attention.

\section{Training and Inference}
\label{sec:training}

The training objective comprises two interrelated tasks: (1) generating high-quality translations at the optimal EMIT moments; (2) learning an online EMIT/WAIT decision policy.

\subsection{Loss Heatmap Construction}

Given an input sequence of length $X$ and a target sequence of length $Y$, the model performs a full grid forward pass and computes the log-probability of each target token at every grid state:
\begin{equation}
  \mathcal{L}(x,y) = -\log p[x,y]_{o_{y+1}} = \mathrm{CE}(p[x,y],\; o_{y+1}),
  \label{eq:grid-loss}
\end{equation}
yielding a \textbf{loss heatmap} $\mathcal{L} \in \mathbb{R}^{X \times Y}$, where a smaller value at $(x,y)$ indicates that the model can predict $o_{y+1}$ more accurately given $x$ input tokens and $y$ output tokens.

The heatmap encodes all the information required to make optimal EMIT/WAIT decisions. For a fixed output position $y$, observing $\mathcal{L}(x,y)$ as $x$ increases reveals how much each new input token helps:
\begin{itemize}
  \item If $\mathcal{L}(x,y) < \mathcal{L}(x-1,y)$, the new input token $i_x$ provides information useful for predicting $o_{y+1}$, so WAIT is beneficial at this state.
  \item If $\mathcal{L}(x,y) \approx \mathcal{L}(x-1,y)$, the new input contributes little to the current output prediction; WAIT only adds latency, making EMIT the better choice.
\end{itemize}
The optimal simultaneous translation strategy is therefore equivalent to finding a monotone path on the heatmap that minimizes the total EMIT-step loss while maximizing the area below the path (i.e., the latency reward).

\subsection{Optimal Monotone Path via Dynamic Programming}

A \textbf{monotone path} $\pi$ from $(0,0)$ to $(X,Y)$ consists of exactly $X$ WAIT steps and $Y$ EMIT steps. We define the path score as a trade-off between translation quality and latency:
\begin{equation}
  R(\pi) = \lambda \cdot \mathrm{Area}(\pi) - \sum_{\substack{(x,y)\to(x,y+1)\\\in\,\pi}} \mathcal{L}(x,y),
  \label{eq:path-score}
\end{equation}
where $\mathrm{Area}(\pi)$ counts the grid cells below the path (rewarding early WAIT), and $\lambda > 0$ is the latency tolerance hyperparameter. Intuitively, $\lambda$ sets the exchange rate between one unit of path area (one extra WAIT step at height $y$) and one unit of translation loss. A larger $\lambda$ makes area more valuable relative to loss, pushing the optimal path to WAIT longer before emitting and accumulating more input context; translation quality may improve but latency increases. A smaller $\lambda$ makes the path prefer early emission, reducing latency at the potential cost of translating with less context. In our experiments we fix $\lambda = 0.1$, which we found to yield a good balance between quality and latency on the development set.

We find the optimal path via dynamic programming:
\begin{equation}
  \mathrm{dp}[x][y] = \max
  \begin{cases}
    \mathrm{dp}[x{-}1][y] + \lambda y, & x > 0 \\[2pt]
    \mathrm{dp}[x][y{-}1] - \mathcal{L}(x,y{-}1), & y > 0,\ x < X
  \end{cases}
  \label{eq:dp-recur}
\end{equation}
with $\mathrm{dp}[0][0] = 0$. The time and space complexity of the DP is $O(XY)$, matching the grid forward pass.

After the DP table is filled, the optimal path is recovered by backtracking from $(X,Y)$ along parent pointers: if the optimal transition into $(x,y)$ was a WAIT step (from $(x{-}1,y)$), we record the row-exit value $y_{\mathrm{at}}[x{-}1] = y$ and move to $(x{-}1,y)$; if it was an EMIT step (from $(x,y{-}1)$), we move to $(x,y{-}1)$ without updating $y_{\mathrm{at}}$. This yields the row-exit sequence $y_{\mathrm{at}}[0],\ldots,y_{\mathrm{at}}[X{-}1]$, where $y_{\mathrm{at}}[x]$ is the $y$-coordinate at which the path leaves row $x$ (i.e., the number of EMIT steps performed before the $(x{+}1)$-th WAIT).

\begin{figure}[htbp]
  \centering
  \includegraphics[width=0.78\linewidth]{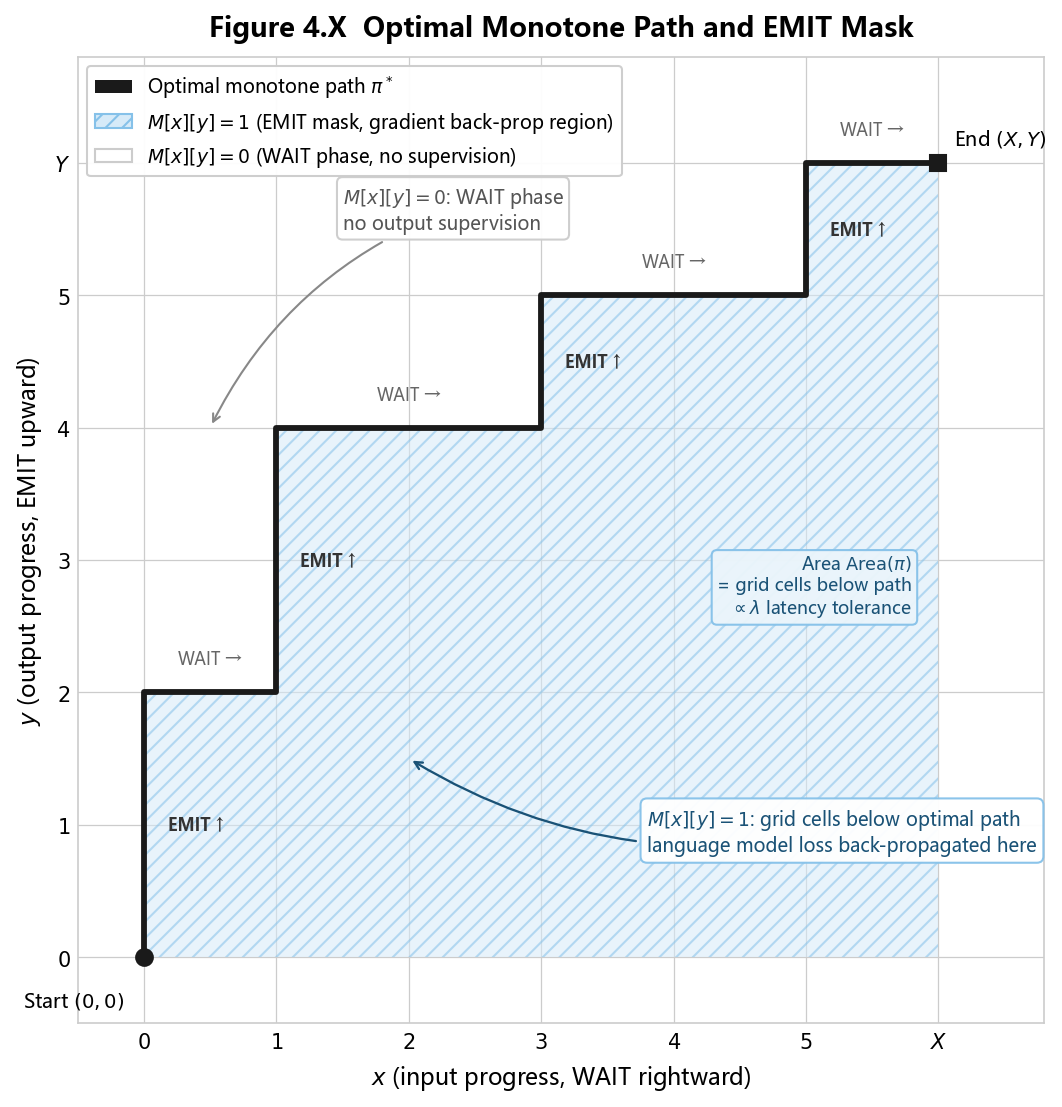}
  \caption{Illustration of the loss heatmap and the optimal DP path. Color intensity indicates the grid loss $\mathcal{L}(x,y)$ (darker = higher loss). The black staircase line is the optimal monotone path: horizontal segments are WAIT steps and vertical segments are EMIT steps.}
  \label{fig:dp-path}
\end{figure}

\subsection{EMIT Mask and Joint Training Loss}

From the optimal path's row-exit sequence $\{y_{\mathrm{at}}[x]\}$, we derive an \textbf{EMIT mask} $M \in \{0,1\}^{X \times Y}$:
\begin{equation}
  M[x][y] = \mathbf{1}[y < y_{\mathrm{at}}[x]].
  \label{eq:emit-mask}
\end{equation}
Geometrically, $M[x][y] = 1$ means that at state $(x,y)$ the optimal path has already performed at least $y$ EMIT steps while on row $x$, i.e., the path lies above position $y$ on that row. The set of cells with $M[x][y]=1$ therefore forms exactly the region below the optimal path, so $\sum_{x,y} M[x][y] = \mathrm{Area}(\pi^*)$---directly connecting the mask to the area reward in the path score.

The language model loss is computed only on EMIT-masked cells:
\begin{equation}
  \mathcal{L}_{\mathrm{LM}} = \frac{1}{|M_+|} \sum_{(x,y):\, M[x][y]=1} \mathcal{L}(x,y).
  \label{eq:lm-loss}
\end{equation}
A binary EMIT decision head is trained with cross-entropy against the mask:
\begin{equation}
  \mathcal{L}_{\mathrm{emit}} = \frac{1}{|\mathcal{V}|} \sum_{(x,y) \in \mathcal{V}} \mathrm{CE}(e[x][y],\; M[x][y]).
  \label{eq:emit-loss}
\end{equation}
The total training loss is:
\begin{equation}
  \mathcal{L}_{\mathrm{total}} = \mathcal{L}_{\mathrm{LM}} + \mathcal{L}_{\mathrm{emit}}.
  \label{eq:total-loss}
\end{equation}
The two losses have complementary roles: $\mathcal{L}_{\mathrm{LM}}$ trains translation ability, ensuring that tokens emitted at the optimal moment are semantically accurate; $\mathcal{L}_{\mathrm{emit}}$ trains the decision policy, teaching the EMIT head to reproduce the optimal path during inference. Together they form a \textbf{self-guided online path optimization} loop that requires no external alignment annotation:
\begin{enumerate}
  \item Improved translation ability lowers grid losses $\mathcal{L}(x,y)$, reshaping the loss heatmap.
  \item A reshaped heatmap yields a better optimal path via DP, providing higher-quality supervision labels for the EMIT head.
  \item A better-trained EMIT head makes inference-time decisions closer to the optimal path, supplying the language model with more appropriate input context.
  \item More appropriate context in turn further improves translation quality, closing the loop.
\end{enumerate}
Through this cycle, the model self-consistently learns the optimal latency-quality trade-off as translation competence and scheduling competence co-evolve throughout training.

\subsection{Incremental KV Cache Design}
\label{sec:inference}

Naively re-running the full $X \times Y$ grid forward pass at each step would cost $O(XYL)$ per step. We design an incremental KV cache that reduces EMIT steps to $O(X_{\mathrm{vis}} \cdot L)$ and WAIT steps to $O(Y_{\mathrm{gen}} \cdot L)$.

The cache stores, per layer, eight types of tensors: the keys and values for I$\to$I and O$\to$O (under broadcast approximation), their log-sum-exp statistics for joint Softmax, and the running prefix statistics $(m, Z, S)$ for the two cross-attention types (I$\leftarrow$O and O$\leftarrow$I). Total cache memory is approximately:
\begin{multline}
  \mathcal{M}_{\mathrm{cache}} = L \cdot [2(X+Y) H_k H_d \\
  + 6(X+Y) H + 2(X+Y) H H_d] \cdot 2\,\text{bytes (fp16)},
\end{multline}
which is around 27 MB for $L=24$, $X=Y=128$, $H=14$, $H_k=2$, $H_d=64$---far smaller than the model parameters.

An \textbf{EMIT step} computes only the new O-stream column $y_{\mathrm{new}}$ for all $x \in \{0,\ldots,x_{\mathrm{vis}}-1\}$: it reads $\mathbf{K}_O^{x=0}$, $\mathbf{V}_O^{x=0}$ from cache for O$\to$O, uses cached I-stream keys and prefix statistics for O$\leftarrow$I, and updates the I$\leftarrow$O prefix statistics with the new O-stream hidden state. A \textbf{WAIT step} is symmetric: it computes the new I-stream row $x_{\mathrm{new}}$ and updates O$\leftarrow$I prefix statistics.

\subsection{KV Cache Structure}
The duplex inference cache maintains eight categories of tensors per layer. Table~\ref{tab:kvcache-entries} summarizes the cached items and their update rules.

\begin{table}[htbp]
  \caption{Per-layer KV cache entries and update rules}
  \label{tab:kvcache-entries}
  \centering
  \begin{tabular}{llll}
    \toprule
    Cache Entry & Purpose & Update Step \\
    \midrule
    $\mathbf{K}_I^{y=0}$ & I$\to$I broadcast keys & WAIT \\
    $\mathbf{V}_I^{y=0}$ & I$\to$I broadcast values & WAIT \\
    $\mathbf{K}_O^{x=0}$ & O$\to$O broadcast keys & EMIT \\
    $\mathbf{V}_O^{x=0}$ & O$\to$O broadcast values & EMIT \\
    $m^{\mathrm{ii}}, Z^{\mathrm{ii}}, S^{\mathrm{ii}}$ & I$\to$I joint Softmax statistics & WAIT \\
    $m^{\mathrm{oo}}, Z^{\mathrm{oo}}, S^{\mathrm{oo}}$ & O$\to$O joint Softmax statistics & EMIT \\
    $m^{\mathrm{io}}, Z^{\mathrm{io}}, S^{\mathrm{io}}$ & I$\leftarrow$O prefix statistics & EMIT \\
    $m^{\mathrm{oi}}, Z^{\mathrm{oi}}, S^{\mathrm{oi}}$ & O$\leftarrow$I prefix statistics & WAIT \\
    \bottomrule
  \end{tabular}
\end{table}

Overall cache memory is approximately:
\begin{multline}
  \mathcal{M}_{\mathrm{cache}} = L \cdot \bigl[2(X+Y) H_k H_d \\
  + 6(X+Y) H + 2(X+Y) H H_d\bigr] \cdot 2\,\text{bytes (fp16)}.
\end{multline}

With $L = 24$, $X = Y = 128$, $H = 14$, $H_k = 2$, $H_d = 64$, the cache requires roughly 27 MB, which is modest relative to the full model footprint.

\subsection{Incremental Inference Algorithm}
The full duplex inference loop alternates EMIT and WAIT steps, reusing cached keys, values, and prefix statistics to avoid repeated full-grid recomputation. Table~\ref{tab:duplex-inference} summarizes the main loop.

\begin{table}[htbp]
  \caption{Pseudocode for the duplex inference main loop}
  \label{tab:duplex-inference}
  \centering
  \begin{tabular}{p{1.0\linewidth}}
    \toprule
    Initialize cache with $i_{1:1}$ and optional seed output $o_{1:Y_{\mathrm{seed}}}$. Set $x_{\mathrm{vis}} = 1$, $y_{\mathrm{gen}} = Y_{\mathrm{seed}}$. \\
    \midrule
    While not terminated: \\
    \,\, Compute new O-stream column for current $y_{\mathrm{gen}}$ using cached I$\leftarrow$O and O$\to$O statistics. \\
    \,\, Sample next token $o_{y_{\mathrm{gen}}+1}$. \\
    \,\, If EMIT head indicates WAIT and $x_{\mathrm{vis}} < X_{\mathrm{total}}$, read next input token and update I-stream row cache. \\
    \,\, Otherwise continue EMIT. \\
    \midrule
    Return generated output sequence. \\
    \bottomrule
  \end{tabular}
\end{table}

\begin{figure}[htbp]
  \centering
  \includegraphics[width=1.0\linewidth]{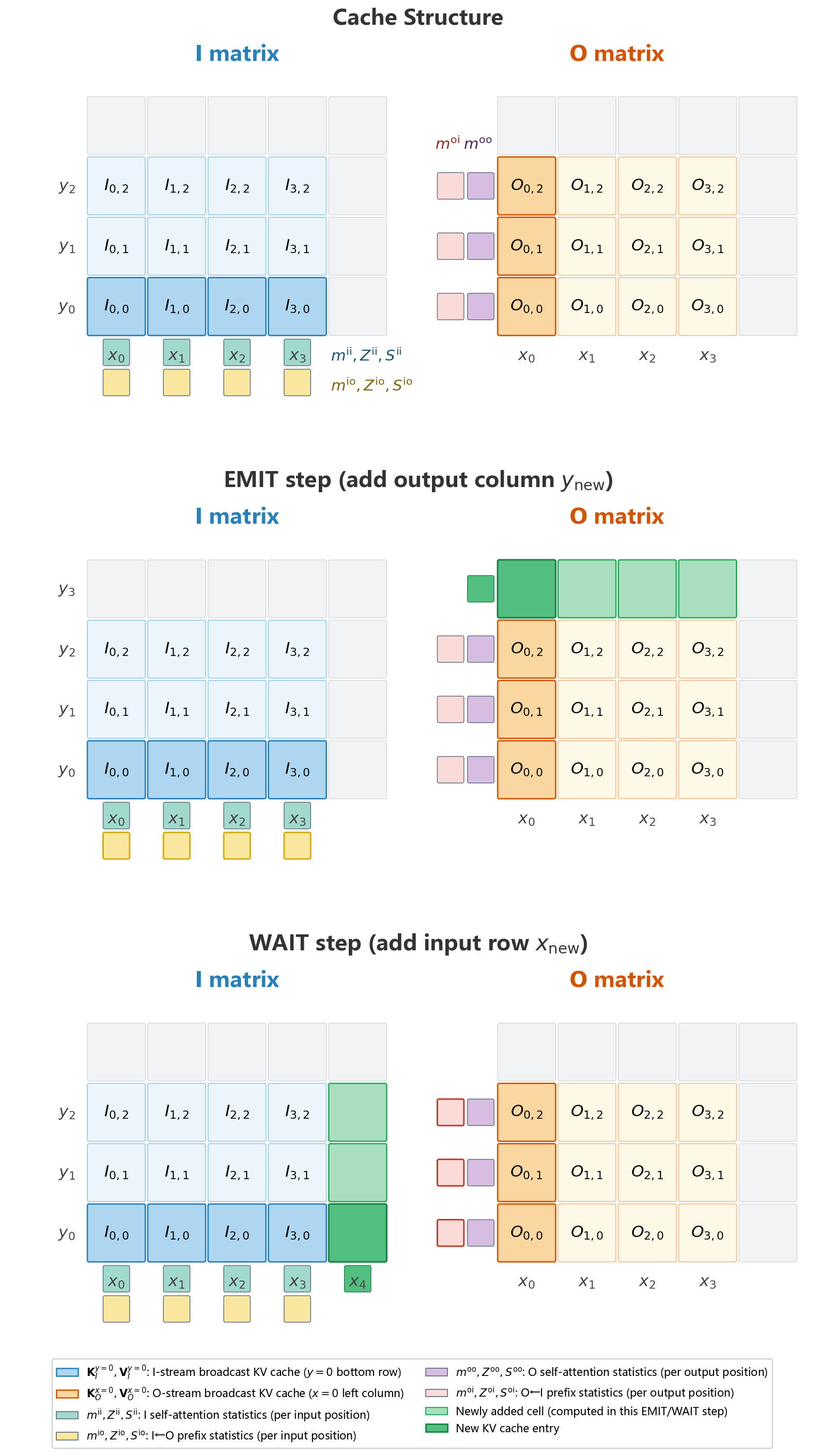}
  \caption{Incremental KV cache update. Left: EMIT step appends a new O-stream column and updates cross-stream statistics. Right: WAIT step appends a new I-stream row. Gray regions are cached; colored regions are newly computed.}
  \label{fig:kvcache-update}
\end{figure}

\subsection{RoPE Anchoring Strategy}

A naive approach assigns output token at position $y$ the RoPE index $x_{\mathrm{vis}} + y$. However, each WAIT step increments $x_{\mathrm{vis}}$, invalidating all cached O-stream keys and requiring $O(Y_{\mathrm{gen}} \cdot L)$ recomputation---defeating the purpose of caching.

We propose an \textbf{anchored RoPE} \cite{su2024roformer} strategy: fix the output position encoding to the \textit{total} input length $X_{\mathrm{total}}$ (known or estimated at inference start):
\begin{equation}
  \mathrm{pos}_O(y) = X_{\mathrm{total}} + y.
  \label{eq:rope-anchor}
\end{equation}
Since $X_{\mathrm{total}}$ is constant throughout inference, all cached O-stream key vectors remain valid across WAIT steps. The resulting full-sequence inference complexity is $O((X^2 + XY + Y^2) \cdot L)$, a reduction of $\min(X,Y)\times$ over the naive approach.

\section{Experiments}
\label{sec:experiments}

\subsection{Experimental Setup}
\label{sec:exp-setup}

\subsubsection{Datasets}

The experiments employ the Chinese-to-English (Zh-En) subset of the WMT 2021 \cite{specia2021findings} news translation task as training data. The dataset combines multiple sources including ParaCrawl \cite{banon2020paracrawl}, News-Commentary, Wiki-Titles, UN Parallel Corpus \cite{ziemski2016united}, WikiMatrix \cite{schwenk2021wikimatrix}, and CCMT \cite{yang2019ccmt}, totaling approximately 25 million bilingual sentence pairs.

Data cleaning operations include:
\begin{itemize}
  \item \textbf{Filtering numeric sequences}: Samples containing meaningless Arabic numerals are removed.
  \item \textbf{Normalizing special punctuation}: MyBatis-style escaped punctuation (e.g., \texttt{\&apos;}, \texttt{\&quot;}) are mapped to standard punctuation.
  \item \textbf{Space normalization}: All spaces are removed from Chinese text; redundant spaces before punctuation are removed from English text.
\end{itemize}

After cleaning, approximately 16 million valid sentence pairs remain. To control GPU memory during training, data are uniformly sampled such that both source and target sequences have length $\le 32$. This yields 100K training examples and 1K validation examples.

\subsubsection{Evaluation Metrics and Baselines}

\begin{itemize}
    \item \textbf{Translation Quality.}
We evaluate translation quality using BLEURT~\cite{papineni2002bleu} and COMET \cite{rei2020comet}. BLEURT is a learned reference-based metric that fine-tunes a BERT-like encoder on human judgement data; it captures semantic equivalence and fluency beyond surface n-gram overlap. COMET (Crosslingual Optimized Metric for Evaluation of Translation) is a neural metric trained on direct assessment scores from professional translators, and has shown strong correlation with human judgements across language pairs in recent WMT shared tasks.
    \item \textbf{Latency.}
We adopt three complementary latency metrics. \textit{Average Lagging} (AL)~\cite{elbayad2020efficient} measures the average number of extra source tokens the system has consumed relative to an ideal simultaneous policy that reads and writes at the same rate; lower AL indicates tighter source--target synchronization. \textit{Average Proportion} (AP) measures the average fraction of the source sentence consumed at the time each output token is emitted; AP\,=\,1.00 means the full input was read before any output was produced. \textit{First Response Latency} (FRL) is the number of input tokens consumed before the first output token is generated, reflecting how quickly the system begins responding; for the grid model, this corresponds to the $x$-coordinate of the first EMIT state.
    \item \textbf{Baselines.}
We compare against two systems. The \textit{non-streaming baseline} is the unmodified Qwen2.5-0.5B model prompted to translate after receiving the complete source input (AP\,=\,1.00); it represents the quality upper bound with no latency constraint. The \textit{Wait-$k$ baseline} couples the same Qwen2.5-0.5B \cite{yang2025qwen3} backbone with the Wait-$k$ policy, evaluated at $k \in \{3, 5\}$; it represents the standard fixed-policy simultaneous translation approach.
\end{itemize}

\subsubsection{Training Details}

Experiments are conducted on 4 NVIDIA RTX 3090 GPUs. The Qwen2.5-0.5B \cite{yang2025qwen3} backbone uses LoRA \cite{hu2021lora} fine-tuning; the text decoder head and EMIT decision head use full fine-tuning. Hyperparameters are shown in Table~\ref{tab:train-hparam}.

\begin{table}[htbp]
  \centering
  \caption{Dual-stream model training hyperparameters}
  \label{tab:train-hparam}
  \begin{tabular}{cc}
    \toprule
    \textbf{Hyperparameter} & \textbf{Value} \\
    \midrule
    Base model & Qwen2.5-0.5B \\
    LoRA rank / scale & 64 / 128 \\
    LoRA modules & \makecell{q\_proj, k\_proj, v\_proj, o\_proj,\\gate\_proj, up\_proj, down\_proj} \\
    Text decoder & Full fine-tuning \\
    EMIT decision head & Full fine-tuning \\
    Learning rate & $1\times10^{-4}$ \\
    Batch size per GPU & 4 \\
    Precision & fp16 \\
    Optimizer & Adam \\
    Warmup steps & 3000 \\
    Training epochs & 1 \\
    Path hyperparameter $\lambda$ & 0.1 \\
    Hardware & 4$\times$NVIDIA RTX 3090 \\
    \bottomrule
  \end{tabular}
\end{table}

Training uses fp16 mixed precision with Adam \cite{Kingma2014AdamAM} optimizer, initial learning rate $1\times10^{-4}$, and 3000 warmup steps. All linear projection layers in the backbone are adapted with LoRA (rank 64, scale 128); decoder and EMIT heads use full fine-tuning. The dynamic programming optimal path is recomputed online per minibatch; latency tolerance hyperparameter $\lambda$ is fixed at 0.1.

\subsection{Experimental Results}
\label{sec:exp-results}

\subsubsection{Case Study: Clause Reordering}

To qualitatively illustrate the proposed model's translation and decision behavior, we analyze a Chinese-to-English translation pair containing a relative clause:
\begin{table}[htbp]
  \centering
  \caption{Chinese-to-English translation case study}
  \label{tab:case-study}
  \begin{tabular}{ll}
    \toprule
    \textbf{Source} & 站在那边的男人曾是棒球手。\\
    \textbf{Reference} & The man standing there was a baseball player. \\
    \bottomrule
  \end{tabular}
\end{table}

The core difficulty is \textbf{word order inversion between Chinese and English relative clauses}: the Chinese modifier (``standing there'') precedes the head noun (``man''), while English requires the head noun ``The man'' to appear first, followed by the modifier ``standing there''. The model must wait for the head noun to appear before outputting in the correct English word order.

\paragraph{Grid Forward Propagation Visualization}

\begin{figure}[htbp]
  \centering
  \includegraphics[width=0.85\linewidth]{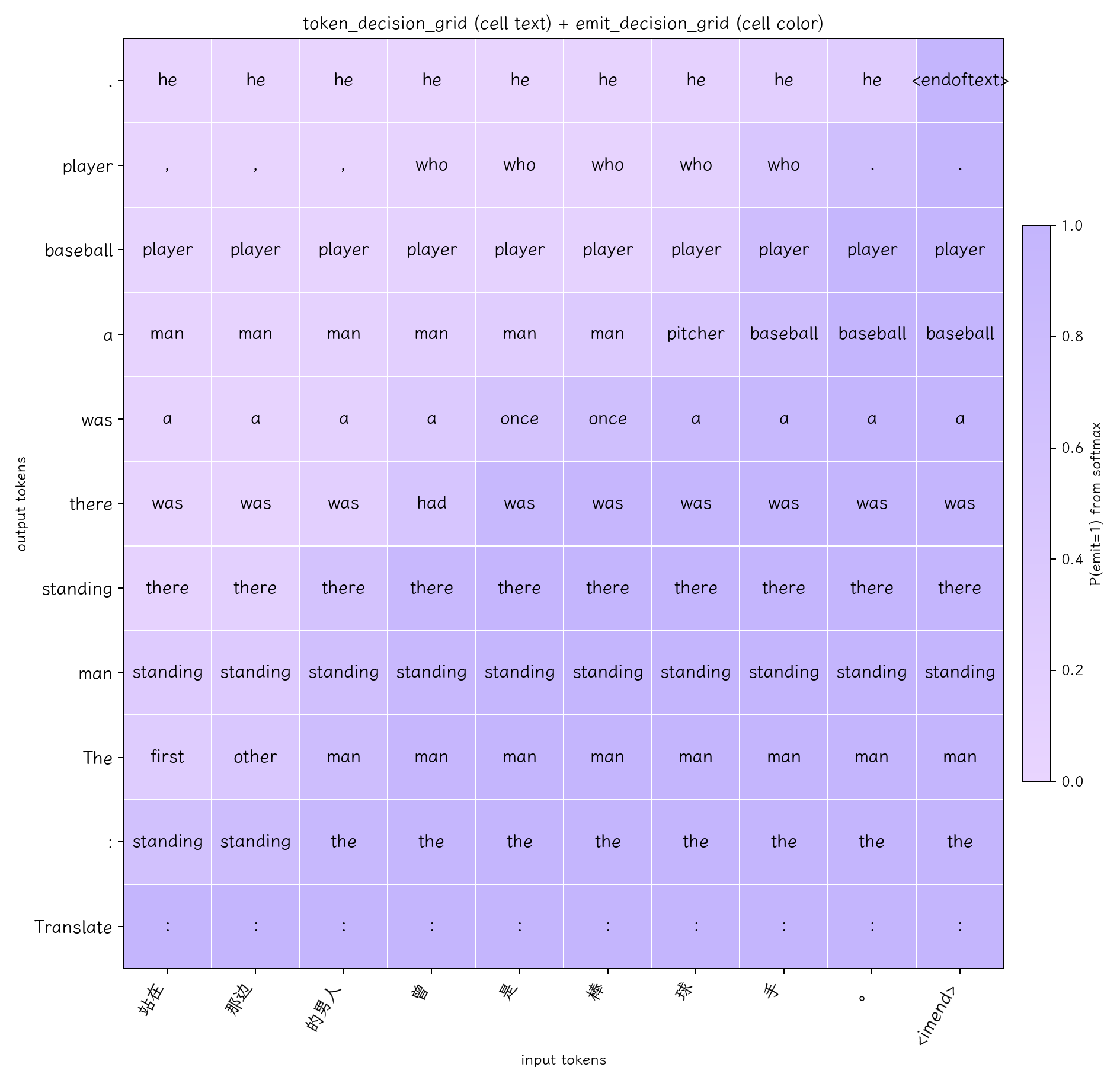}
  \caption{Grid forward propagation visualization. Horizontal axis: input sequence (left to right). Vertical axis: output sequence (bottom to top). Cell text: highest-probability token. Cell color: EMIT probability intensity.}
  \label{fig:grid-output}
\end{figure}

Figure~\ref{fig:grid-output} visualizes the grid forward pass for the above source and reference as I-stream and O-stream inputs. Each cell contains the highest-probability predicted token; background color intensity indicates EMIT decision confidence (darker indicates stronger EMIT signal).

Key observations:
\begin{enumerate}
  \item \textbf{Overall trend}: EMIT probability exhibits a left-to-right, bottom-to-top gradation (lighter $\to$ darker). Longer input context and shorter pending output correlate with stronger EMIT confidence.
  \item \textbf{Clause region response}: When reading input token~1 (``standing''), the predicted output is ``standing'' but EMIT probability is low. The model recognizes the semantic correspondence but refrains from output because English word order requires the head noun first.
  \item \textbf{Post-head-noun transition}: Upon reading input token~3 (``the man''), the prediction shifts to ``the'' and EMIT probability increases significantly. The model correctly identifies that the head noun has appeared and begins output in proper English order.
\end{enumerate}

This demonstrates that the model has internalized Chinese-English clause reordering at the attention level: ``know the translation but hold back'' during modifiers, then ``switch target and commit'' once the head noun appears.

\paragraph{Inference Process Under Different EMIT Thresholds}

Table~\ref{tab:infer-process} shows simultaneous inference with varying EMIT thresholds $\theta$ (emission only when predicted EMIT probability exceeds $\theta$).

\begin{table*}[htbp]
  \centering
  \caption{Simultaneous inference under different EMIT thresholds}
  \label{tab:infer-process}
  \resizebox{\textwidth}{!}{\begin{tabular}{c|cccccccccc}
    \toprule
    $\theta$ & 站在 & 那边 & 的男人 & 曾 & 是 & 棒 & 球 & 手 & 。 & \texttt{<eos>} \\
    \midrule
    0.3 & standing & & there, the man & had & been & a & pitcher & & & . \\
    0.4 & & the & & man standing there had & been & a & pitcher & & & . \\
    0.5 & & & the & man standing there & was & a & & baseball & player & . \\
    0.6 & & & & the man standing there & was & a & & baseball & player & . \\
    0.7 & & & & the man standing there & was & & a & & baseball player & . \\
    0.8 & & & & the & man standing there & was & & a & baseball & player . \\
    0.9 & & & & & the & man standing there & was & & a & baseball player . \\
    \bottomrule
  \end{tabular}}
\end{table*}

\textbf{Low threshold ($\theta = 0.3 \sim 0.4$):} Premature emission causes word order and semantic errors. At $\theta = 0.3$, ``standing'' is output upon reading token~1, then ``there, the man'' at token~3. This reflects Chinese word order (modifier before head), deviating from English. Additionally, ``baseball player'' (tokens 6--8) is incorrectly translated to ``pitcher'' and ``was'' (tokens 4--5) to ``had been'', showing that insufficient context harms both syntax and semantics.

\textbf{Medium threshold ($\theta = 0.5 \sim 0.7$):} Correct waiting followed by quality output. At $\theta = 0.5$, the model WAITs at tokens~1 and~2, then outputs ``the'' upon token~3, followed by ``man standing there'' at token~4. This achieves perfect clause reordering. $\theta = 0.6 \sim 0.7$ behave similarly, with output aligned to the reference: ``was a baseball player''. Sufficient context ensures both grammatical and semantic correctness.

\textbf{High threshold ($\theta = 0.8 \sim 0.9$):} Correct translation but unnecessarily high latency. At $\theta = 0.8$, output starts at token~4 (FRL\,=\,4); at $\theta = 0.9$, at token~5 (FRL\,=\,5). Translation quality matches medium thresholds but First Response Latency increases from 3 to 5, introducing avoidable delay.

\subsubsection{Quantitative Results}

Table~\ref{tab:duplex-results} reports BLEURT, COMET, Average Lagging (AL), Average Proportion (AP), and First Response Latency (FRL) for the proposed model across five EMIT thresholds, the Wait-$k$ baseline at $k \in \{3, 5\}$, and the baseline (Qwen2.5-0.5B translating after the full input is received), which serves as a quality reference upper bound.

\begin{table}[htbp]
  \centering
  \caption{Translation quality and latency on the zh$\to$en test set. The baseline reads the complete input before translating ($\mathrm{AP}=1.00$).
           Bold indicates the best value in each column.}
  \label{tab:duplex-results}
  \resizebox{\linewidth}{!}{\begin{tabular}{llccccc}
    \toprule
    \textbf{System} & \textbf{Param}
      & \textbf{BLEURT}$\uparrow$
      & \textbf{COMET}$\uparrow$
      & \textbf{AL}$\downarrow$
      & \textbf{AP}$\downarrow$
      & \textbf{FRL}$\downarrow$ \\
    \midrule
    Base model & ---
      & \textbf{54.03} & 63.85 & 9.39 & 1.00 & 18.06 \\
    \midrule
    \multirow{2}{*}{Wait-$k$}
      & $k=3$ & 43.20 & 52.06 & 4.98 & 0.79 & 3.00 \\
      & $k=5$ & 45.20 & 54.93 & 5.99 & 0.84 & 4.96 \\
    \midrule
    \multirow{5}{*}{Ours}
      & $\theta=0.5$ & 47.01 & 59.59 & \textbf{0.23} & \textbf{0.49} & \textbf{2.76} \\
      & $\theta=0.6$ & 48.86 & 62.42 & 1.14 & 0.53 & 3.31 \\
      & $\theta=0.7$ & 49.10 & 62.79 & 1.61 & 0.55 & 3.79 \\
      & $\theta=0.8$ & 49.75 & 63.54 & 2.87 & 0.59 & 4.65 \\
      & $\theta=0.9$ & 50.74 & \textbf{64.73} & 4.39 & 0.65 & 6.09 \\
    \bottomrule
  \end{tabular}
  }
\end{table}

\begin{figure}[htbp]
  \centering
  \includegraphics[width=1.0\linewidth]{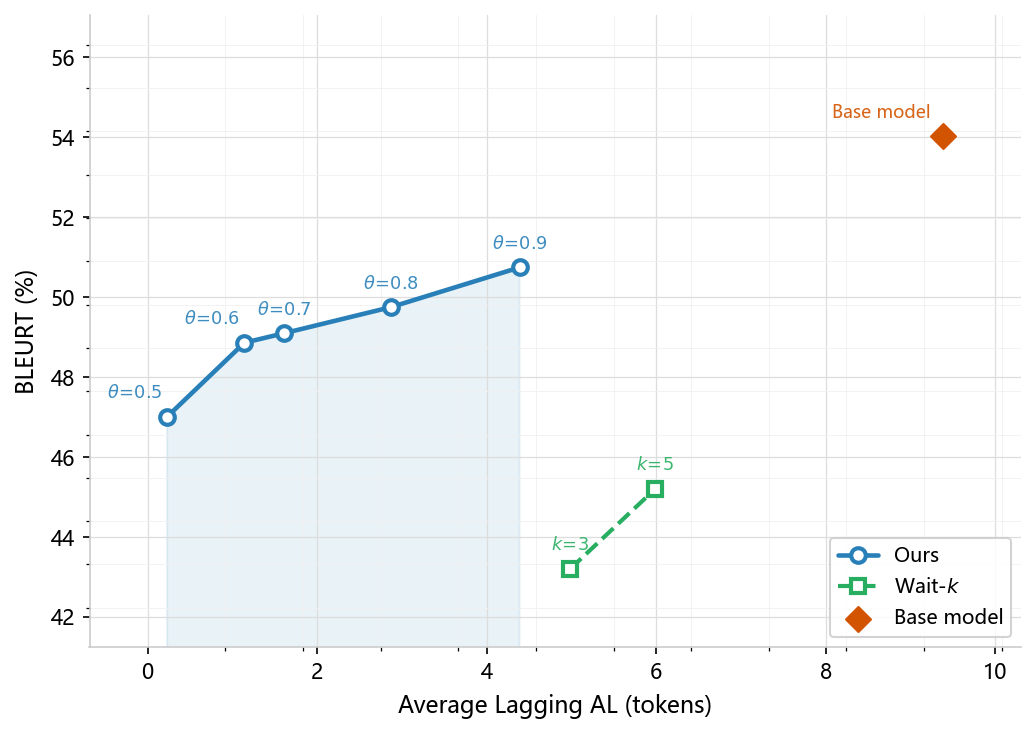}
  \caption{BLEURT--AL quality--latency trade-off on the zh$\to$en test set.}
  \label{fig:al-bleurt}
\end{figure}

\textbf{Quality--latency trade-off.}
Within the proposed model, raising $\theta$ from 0.5 to 0.9 monotonically improves translation quality (BLEURT: $47.01 \to 50.74$; COMET: $59.59 \to 64.73$) at the cost of increased latency (AL: $0.23 \to 4.39$; FRL: $2.76 \to 6.09$). The emission threshold therefore provides a single, continuous knob for balancing quality against latency without retraining. Notably, at $\theta=0.9$ the proposed model achieves COMET\,=\,64.73, surpassing the base model (63.85) while reducing FRL from 18.06 to 6.09---demonstrating that the dual-stream attention mechanism not only preserves but can exceed offline translation quality at a fraction of the latency. Figure~\ref{fig:al-bleurt} visualizes the AL--BLEURT Pareto frontier for all systems, showing that the proposed model dominates the Wait-$k$ baseline across the full latency range.

\textbf{Comparison with Wait-$k$.}
At comparable first-response latency, the proposed model consistently outperforms Wait-$k$ on both quality and latency efficiency. At FRL $\approx 3$, our model at $\theta=0.6$ (FRL\,=\,3.31) achieves BLEURT\,=\,48.86 and COMET\,=\,62.42, compared with BLEURT\,=\,43.20 and COMET\,=\,52.06 for Wait-$k$ with $k=3$ (FRL\,=\,3.00)---a gain of $+5.66$ BLEURT and $+10.36$ COMET points. At FRL $\approx 5$, our model at $\theta=0.9$ (FRL\,=\,6.09) similarly surpasses Wait-$k$ with $k=5$ (FRL\,=\,4.96) by $+5.54$ BLEURT and $+9.80$ COMET.

Furthermore, the proposed model exhibits substantially lower AL and AP across all operating points. Even at $\theta=0.9$, our AL (4.39) remains below that of Wait-$k$ with $k=3$ (4.98), and our AP (0.49--0.65) is markedly lower than the Wait-$k$ range (0.79--0.84). This confirms that the learned EMIT/WAIT policy avoids the rigid read-ahead imposed by a fixed-$k$ schedule, achieving tighter source--target synchronization while maintaining higher translation quality.

\section{Conclusion}
\label{sec:conclusion}

We presented a dual-stream simultaneous translation model that directly encodes the bidirectional dependency between the input and output streams inside the Transformer attention mechanism, rather than imposing a read-write schedule as post-processing. The key architectural contribution is a two-dimensional grid hidden-state representation in which each cell $(x,y)$ maintains separate input-stream and output-stream hidden states conditioned on $x$ visible source tokens and $y$ generated target tokens. Within this grid, four types of attention---I$\to$I, O$\to$O, I$\leftarrow$O, and O$\leftarrow$I---capture all relevant intra- and cross-stream dependencies under the appropriate causal constraints, and are merged via a joint QK Softmax normalization that lets the model learn the relative importance of self-attention and cross-attention in a data-driven manner.

To make the grid computation tractable, we proposed two complementary approximations: a broadcast approximation that reduces intra-stream self-attention from $O(X^2Y + XY^2)$ to $O(X^2 + Y^2)$, and a Hadamard approximation that reduces cross-attention from $O(XY^2 + X^2Y)$ to $O(XY)$. Both approximations are exact at initialization and are linguistically well-motivated---intra-stream attention patterns and cross-stream attention scores depend primarily on token content and relative position rather than on the precise count of tokens in the other stream.

For training, we constructed a per-grid-cell loss heatmap and found the optimal EMIT/WAIT path via dynamic programming, balancing translation quality against latency through a scalar hyperparameter $\lambda$. The resulting self-guided optimization loop requires no external alignment annotation: improved translation quality reshapes the heatmap, yielding better optimal paths that in turn provide higher-quality supervision for the EMIT decision head, whose improved inference-time decisions further benefit the language model. For inference, an incremental KV cache together with an anchored RoPE strategy reduces the per-step cost from $O(XYL)$ to $O(X_{\mathrm{vis}} L)$ for EMIT steps and $O(Y_{\mathrm{gen}} L)$ for WAIT steps.

Experiments on Chinese-to-English simultaneous translation show that the proposed model substantially outperforms the Wait-$k$ baseline at comparable first-response latency. At FRL\,$\approx$\,3, our model gains $+5.66$ BLEURT and $+10.36$ COMET over Wait-$k$ ($k=3$), while also achieving markedly lower Average Lagging and Average Proportion. At the high-quality end ($\theta=0.9$), the model attains COMET\,=\,64.73, exceeding even the non-streaming baseline (63.85) while reducing first-response latency from 18.06 to 6.09 tokens. These results demonstrate that explicit dual-stream attention is a principled and effective alternative to post-hoc read-write policies for simultaneous translation.

Future work includes scaling to larger backbone models and longer sequences, extending the grid framework to speech inputs and outputs, and exploring differentiable path optimization to allow end-to-end gradient flow through the DP step.

\bibliographystyle{IEEEtran}
\bibliography{ref}

\end{CJK*}
\end{document}